\pdfoutput=1
\documentclass[10pt,twocolumn,letterpaper]{article}

\usepackage{cvpr}
\usepackage{times}
\usepackage{epsfig}
\usepackage{graphicx}
\usepackage{amsmath}
\usepackage{amssymb}
\usepackage{mathrsfs}
\usepackage{array}
\usepackage{booktabs}
\usepackage{adjustbox}
\usepackage{caption}
\usepackage{cuted}
\usepackage{capt-of}
\usepackage{xcolor}
\usepackage{color, colortbl}

\usepackage{floatrow}
\DeclareMathOperator*{\argmin}{arg\,min}

\newcommand\mycom[2]{\genfrac{}{}{0pt}{}{#1}{#2}}
\usepackage[pagebackref=true,breaklinks=true,letterpaper=true,colorlinks,bookmarks=false]{hyperref}

\cvprfinalcopy % *** Uncomment this line for the final submission

\def\cvprPaperID{10193} % *** Enter the CVPR Paper ID here

\definecolor{Gray}{gray}{0.9}

\ifcvprfinal\fi
\begin{document}

%%%%%%%%% TITLE
\title{Unsupervised Intra-domain Adaptation  for Semantic Segmentation \\
through Self-Supervision}

% \author{
% Fei Pan \\
% {\tt\small feipan664@gmail.com} 
% \and Inkyu Shin\\
% {\tt\small dlsrbgg33@kaist.ac.kr}
% \and Francois Rameau \\
% {\tt\small rameau.fr@gmail.com}
% \and Seokju Lee \\
% {\tt\small seokju91@gmail.com}
% \and In So Kweon \\
% {\tt\small iskweon77@kaist.ac.kr} \\
% KAIST, Daejeon, South Korea
% }
\author{
\hspace{0mm}Fei Pan
\hspace{10mm}Inkyu Shin
\hspace{10mm}Francois Rameau
\hspace{10mm}Seokju Lee
\hspace{10mm}In So Kweon
\vspace{1mm}
\\
KAIST, South Korea
\\
\hspace{0mm}{\tt\small \{feipan, dlsrbgg33, frameau, seokju91, iskweon77\}@kaist.ac.kr}
\\
}

\definecolor{lightsalmonpink}{rgb}{1.0, 0.6, 0.6}
\thispagestyle{empty}
\thispagestyle{empty}

\maketitle

%%%%%%%%% ABSTRACT
\begin{abstract}
    Convolutional neural network-based approaches have achieved remarkable progress in semantic segmentation. However, these approaches heavily rely on annotated data which are labor intensive. To cope with this limitation, automatically annotated data generated from graphic engines are used to train segmentation models. However, the models trained from synthetic data are difficult to transfer to real images. To tackle this issue, previous works have considered directly adapting models from the source data to the unlabeled target data (to reduce the inter-domain gap). Nonetheless, these techniques do not consider the large distribution gap among the target data itself (intra-domain gap). In this work, we propose a two-step self-supervised domain adaptation approach to minimize the inter-domain and intra-domain gap together. First, we conduct the inter-domain adaptation of the model; from this adaptation, we separate the target domain into an easy and hard split using an entropy-based ranking function. Finally, to decrease the intra-domain gap, we propose to employ a self-supervised adaptation technique from the easy to the hard split. Experimental results on numerous benchmark datasets highlight the effectiveness of our method against existing state-of-the-art approaches. The source code is available at \url{https://github.com/feipan664/IntraDA.git}.
\end{abstract}

%%%%%%%%% BODY TEXT
\section{Introduction}

\begin{figure}[t]
    \centering
    \includegraphics[width=\textwidth]{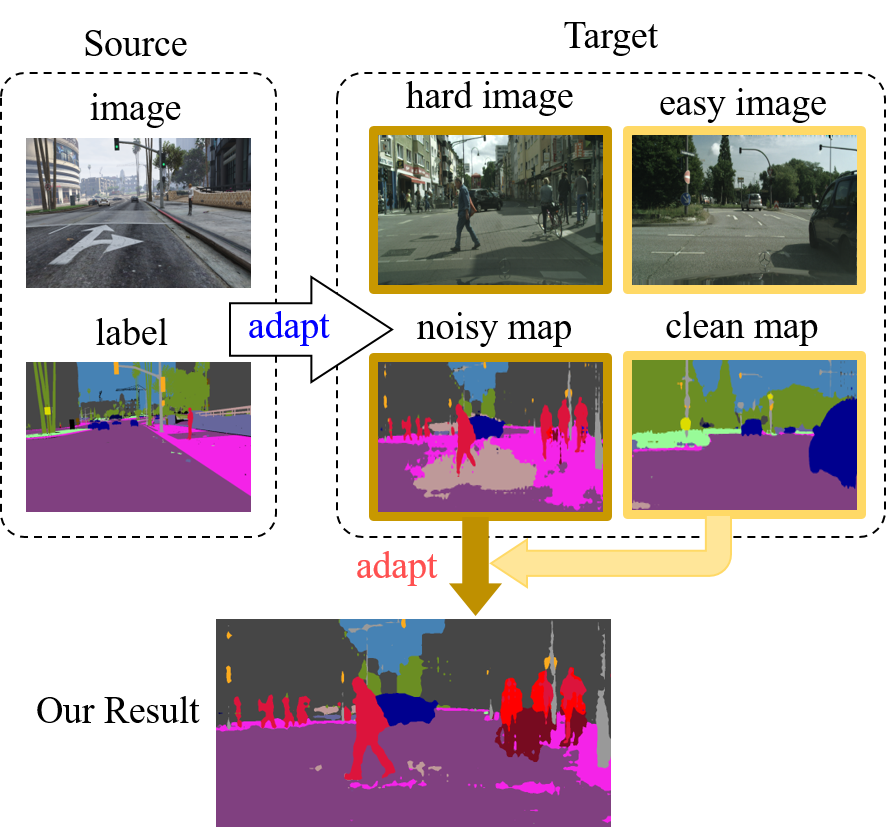}
\caption{We propose a two-step self-supervised domain adaptation technique for semantic segmentation. Previous works solely {\color{blue} adapt} the segmentation model from the source domain to the target domain. Our work also consider {\color{red} adapting} from the clean map to the noisy map within the target domain.}
    \label{fig:introduction}
\end{figure}
%------------------------------------------------------------------------------------------- 
Semantic segmentation aims at assigning each pixel in the image to a semantic class. 
% Convolutional neural network-based segmentation models~\cite{long2015fully, zhao2017pspnet} have achieved remarkable progress with the application in computer vision systems like autonomous driving. 
Recently, convolutional neural network-based segmentation models~\cite{long2015fully, zhao2017pspnet} have achieved remarkable progresses, leading to various applications in computer vision systems, such as autonomous driving~\cite{luc2017predicting,Zhang_2017_ICCV,lee2019visuomotor}, robotics~\cite{milioto2018real,shvets2018automatic}, and disease diagnosis~\cite{zhou2019collaborative,zhao2019data}. 
Training such a segmentation network requires large amounts of annotated data. However, collecting large scale datasets with pixel-level annotations for semantic segmentation is difficult since they are expensive and labor intensive. Recently, photorealistic data rendered from simulators and game engines~\cite{Richter_2016_ECCV, ros2016synthia} with precise pixel-level semantic annotations have been utilized to train segmentation networks. However, the models trained from synthetic data are hardly transferable to real data due to the cross-domain difference~\cite{pmlr-v80-hoffman18a}. To address this issue, unsupervised domain adaptation (UDA) techniques have been proposed to align the distribution shift between the labeled source data and the unlabeled target data. For the particular task of semantic segmentation, adversarial learning-based UDA approaches demonstrate efficiency in aligning features at the image~\cite{murez2018image,pmlr-v80-hoffman18a} or output~\cite{tsai2019domain, tsai2018learning} level. More recently, the entropy of pixel-wise output predictions proposed by~\cite{vu2019advent} is also used for output level alignment. Other approaches~\cite{Zou_2019_ICCV, zou2018unsupervised} involve generating pseudo labels for target data and conducting refinement via an iterative self-training process. While many models consider the single-source-single-target adaptation setting, recent works~\cite{peng2019moment, zhao2018adversarial} have proposed to address the issue of multiple source domains; it focus on the multiple-source-single-target adaptation setting. Above all, previous works have mostly considered adapting models from the source data to the target data (inter-domain gap). 

\begin{figure*}[t!]
    \centering 
    \includegraphics[width=0.98\textwidth]{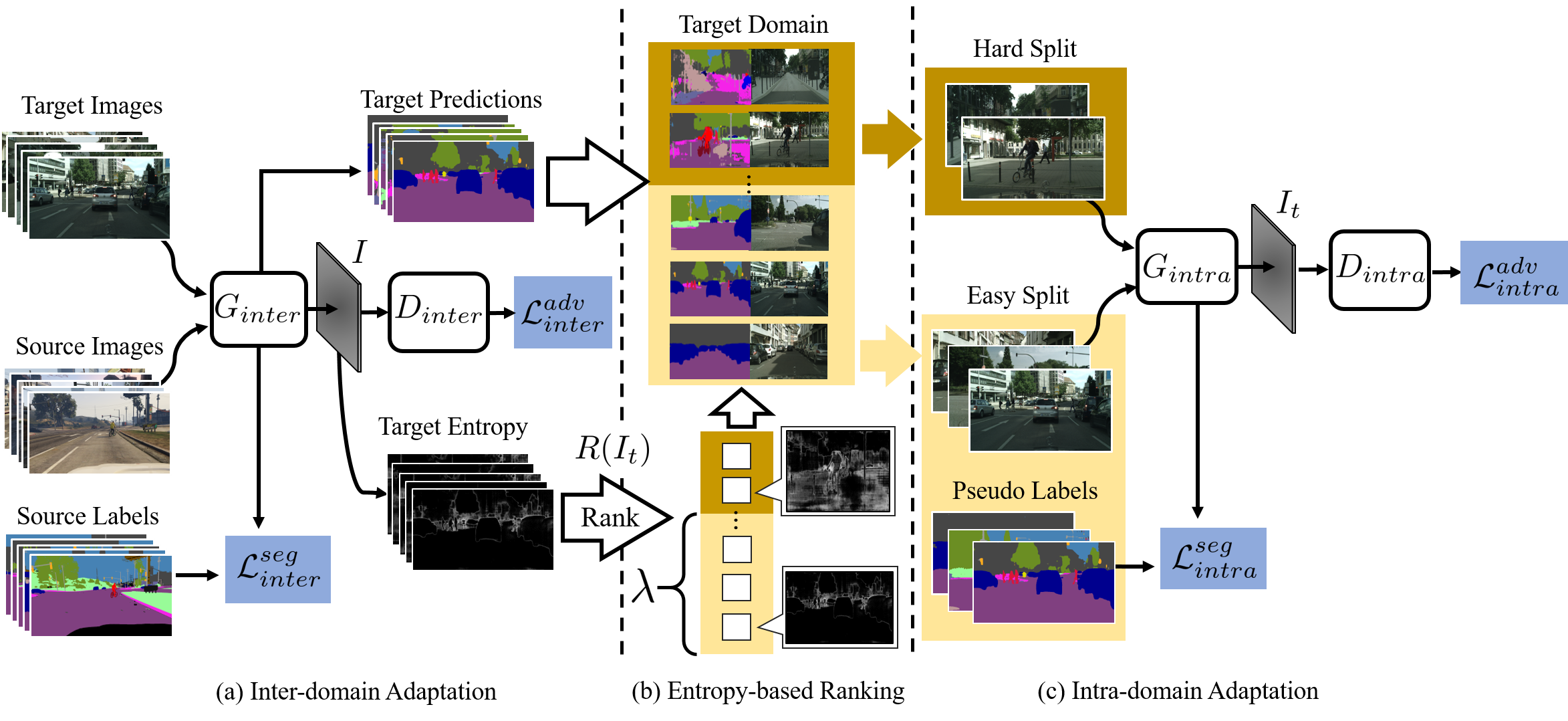}
    \caption{The proposed self-supervised domain adaptation model contains the inter-domain generator and discriminator $\{G_{inter}, D_{inter}\}$, and the intra-domain generator and discriminator $\{G_{intra}, D_{intra}\}$. The proposed model consists of three parts, namely, (a) an inter-domain adaptation, (b) an entropy-based ranking system, and (c) an intra-domain adaptation. In (a), given the source and the unlabeled target data, $D_{inter}$ is trained to predict the domain label for the samples while $G_{inter}$ is trained to fool $D_{inter}$. $\{G_{inter}, D_{inter}\}$ are optimized by minimizing the segmentation loss $\mathcal{L}_{inter}^{seg}$ and the adversarial loss $\mathcal{L}_{inter}^{adv}$. In (b), an entropy based function $R(I_t)$ is used separate all target data into easy split and hard split. An hyperparameter $\lambda$ is introduced as a ratio of target images assigned into the easy split. In (c), an intra-domain adaptation is used to close the gap between easy split and hard split. The segmentation predictions of easy split data from $G_{inter}$ serve as pseudo labels. Given easy split data with pseudo labels and hard split data, $D_{intra}$ is used to predict whether the sample is from easy split or hard split, while $G_{intra}$ is trained to confuse $D_{intra}$. $\{G_{intra}$ and $D_{intra}\}$ are optimized using the intra-domain segmentation loss $\mathcal{L}_{intra}^{seg}$ and the adversarial loss $\mathcal{L}_{intra}^{adv}$.}
    \label{fig:architecture}
\end{figure*}

However, target data collected from the real world have diverse scene distributions; these distributions are caused by various factors such as moving objects, weather conditions, which leads to a large gap in the target (intra-domain gap). For example, the noisy map and clean map in the target domain, shown in Figure~\ref{fig:introduction}, are the predictions made by the same model on different images. While previous studies solely focus on reducing the inter-domain gap, the problem of the intra-domain gap has attracted a relatively low attention. In this paper, we present a two-step domain adaptation approach to minimize the inter-domain and intra-domain gap. Our model consists of three parts, which are presented in Figure~\ref{fig:architecture}, namely, 1) an inter-domain adaptation module to close inter-domain gap between the labeled source data and the unlabeled target data, 2) an entropy-based ranking system to separate target data into the an easy and hard split, and 3) an intra-domain adaptation module to close intra-domain gap between the easy and hard split (using pseudo labels from the easy subdomain). For semantic segmentation, our proposed approach achieves good performance against state-of-the-art approaches on benchmark datasets. Furthermore, our approach outperforms previous domain adaptation approaches for digit classification. \\
 
\noindent \textbf{The Contributions of Our Work.} First, we introduce the inter-domain gap among target data and propose an entropy-based ranking function to separate target domain into an easy and hard subdomain.
Second, we propose a two-step self-supervised domain adaptation approach to minimize the inter-domain and intra-domain gap together. 

% \ik{self-supervsied learning in `inter-domain adaptation' and `intra-domain adaptation'?}

\section{Related Works}
\noindent \textbf{Unsupervised Domain Adaptation.} The goal of unsupervised domain adaptation is to align the distribution shift between the labeled source and the unlabeled target data. Recently, adversarial-based UDA approaches have shown great capabilities in learning domain invariant features, even for complex tasks like semantic segmentation~\cite{vu2019advent,chen2019domain,tsai2019domain,tsai2018learning,saito2018maximum,pmlr-v80-hoffman18a,park2019preserving}. Adversarial-based UDA models for semantic segmentation usually involve two networks. One network is used as a generator to predict the segmentation maps of input images, which can be from the source or the target. Given features from the generator, the second network functions as a discriminator to predict the domain labels. The generator tries to fool the discriminator, so as to align the distribution shift of the features from the two domains. Besides feature level alignment, other approaches try to align domain shift at the image level or output level. At the image level, CycleGAN~\cite{CycleGAN2017} was applied in~\cite{pmlr-v80-hoffman18a} to build generative images for domain alignment. At the output level, \cite{tsai2018learning} proposes an end-to-end model involving structural output alignment for distribution shift. More recently, \cite{vu2019advent} takes advantage of the entropy of pixel-wise predictions from the segmentation outputs to address the domain gap. While all the previous studies exclusively considered aligning the inter-domain gap, our approach further minimizes the intra-domain gap. Thus, our technique can be combined with most existing UDA approaches for extra performance gains. \\

\noindent \textbf{Uncertainty via Entropy.} Uncertainty measurement has a strong connection with unsupervised domain adaptation. For instance,  \cite{vu2019advent} proposes minimizing the target entropy value of the model outputs directly or using adversarial learning~\cite{tsai2018learning,pmlr-v80-hoffman18a} to close the domain gap for semantic segmentation. Also the entropy of the model outputs~\cite{wang2016cost} is used as a confidence measurement for transferring samples across domains~\cite{su2020active}. We propose utilizing entropy to rank target images to separate them into two an easy and hard split. \\

\noindent \textbf{Curriculum Domain Adaptation.} Our work is also related to curriculum domain adaptation~\cite{sakaridis2018model, Zhang_2017_ICCV,dai2019adaptation} which deals with easy samples first. For curriculum domain adaptation on foggy scene understanding, ~\cite{sakaridis2018model} proposes to adapt a semantic segmentation model from non-foggy images to synthetic light foggy images, and then to real heavy foggy images. To generalize this concept, \cite{dai2019adaptation} decomposes the domain discrepancy into multiple smaller discrepancies by introducing unlabeled intermediate domains. However, these techniques require additional information to decompose domains. To cope with this limitation, ~\cite{Zhang_2017_ICCV} focuses on learning the global and local label distributions of images as the first task to regularize the model predictions in the target domain. In contrast, we propose a simpler and data-driven approach to learn the easy target samples based on an entropy ranking system. 

% \ik{What part of this paper is related to curriculum DA? Can I guess that 1) source to whole target 2) easy target to hard target is curriculum way? It needs to be explicitly explained I think. Actually, I think it is not a usual curriculum da setting but another setting of curriculum da like big domain to big domain /  and small subdomain to subdomain.}

%------------------------------------------------ Approach ---------------------------------
% %------------------------------------------------ Figure  -----------------------------------
\section{Approach}
Let $\mathcal{S}$ denote a source domain consisting of a set of images $\subset \mathbb{R}^{H \times W \times 3}$ with their associated ground-truth $C$-class segmentation maps $\subset (1,C)^{H\times W}$; similarly, let $\mathcal{T}$ denote a target domain containing a set of unlabeled images $\subset \mathbb{R}^{H \times W \times 3}$. In this section, a two-step self-supervised domain adaptation for semantic segmentation is introduced. The first step is the inter-domain adaptation, which is based on common UDA approaches~\cite{vu2019advent,tsai2018learning}. Then, the pseudo labels and predicted entropy maps of target data are generated such that the target data can be clustered into an easy and hard split. Specifically, an entropy-based ranking system is used to cluster the target data into the easy and hard split. The second step is the intra-domain adaptation, which consists in aligning the easy split with pseudo labels to the hard split, as shown in Figure~\ref{fig:architecture}. The proposed network consists of the inter-domain generator and discriminator $\{G_{inter}, D_{inter}\}$, and the intra-domain generator and discriminator $\{G_{intra}, D_{intra}\}$. 

%---------------------------------------------------- Inter-domain adaptation --------------------------------------------
\subsection{Inter-domain Adaptation}
A sample $X_s \in \mathbb{R}^{H\times W \times 3}$ is from the source domain with its associated map $Y_s$. Each entry $Y_s^{(h,w)}=\left[ Y_s^{(h,w,c)} \right]_c$ of $Y_s$ provides a label of a pixel $(h,w)$ as a one-hot vector. The network $G_{inter}$ takes $X_s$ as an input and generates a ``soft-segmentation map" $P_s=G_{inter}(X_s)$. Each $C$-dimensional vector $\left[ P_s^{(h,w,c)} \right]_{c}$ at a pixel $(h,w)$ serves as a discrete distribution over $C$ classes. Given $X_s$ with its ground-truth annotation $Y_s$, $G_{inter}$ is optimized in a supervised way by minimizing the cross-entropy loss:  
\begin{equation}
    \mathcal{L}_{inter}^{seg}(X_s, Y_s) = - \sum_{h,w}\sum_{c}Y_s^{(h,w,c)}\log(P_s^{(h,w,c)}).
\end{equation} 
To close the inter-domain gap between the source and target domains, \cite{vu2019advent} proposes to utilize entropy maps in order to align the distribution shift of the features. The assumption of~\cite{vu2019advent} is that the trained models tend to produce over-confident (low-entropy) predictions for source-like images, and under-confident (high-entropy) predictions for target-like images. Due to its simplicity and effectiveness, \cite{vu2019advent} is adopted in our work to conduct inter-domain adaptation. The generator $G_{inter}$ takes a target image $X_t$ as an input and produce the segmentation map $P_t=G_{inter}(X_t)$; the entropy map $I_t$ is formulated as:
\begin{equation}
    I_t^{(h,w)} = \sum_{c} -P_t^{(h,w,c)} \log(P_t^{(h,w,c)}).
\end{equation}
To align the inter-domain gap,  $D_{inter}$ is trained to predict the domain labels for the entropy maps, while $G_{inter}$ is trained to fool $D_{inter}$; the optimization of $G_{inter}$ and $D_{inter}$ is achieved via the following loss function:
\begin{equation}
    \begin{split}
    \mathcal{L}_{inter}^{adv} (X_s, X_t) =  \sum_{h,w}  \log(1 - D_{inter}(I_t^{(h,w)})) & \\
                                +     \log(D_{inter}(I_s^{(h,w)})) &, 
    \end{split}
\end{equation}
where $I_s$ is the entropy map of $X_s$. The loss function $\mathcal{L}_{inter}^{adv}$ and $\mathcal{L}_{inter}^{seg}$ are optimized to align the distribution shift between the source and target data. However, there remains a need for an efficient method that can minimize the intra-domain gap. For this purpose, we propose to separate the target domain into an easy and hard split and to conduct an intra-domain adaptation.  
%---------------------------------------------------Entropy-based ranking -------------------------------------------------

\subsection{Entropy-based Ranking}
Target images collected from the real world have diverse distributions due to various weather conditions, moving objects, and shading. In Figure~\ref{fig:architecture}, some target prediction maps are clean\footnote{The clean prediction map means that the prediction is confident and smooth.} and others are very noisy, despite being generated from the same model. Since the intra-domain gap exists among target images, a straightforward solution is to decompose the target domain into small subdomains/splits. However, it remains a challenging task due to the lack of target labels. To build these splits, we take advantage of the entropy maps in order to determine the confidence levels of the target predictions. The generator $G_{inter}$ takes a target image $X_t$ as input to generate $P_t$ and the entropy map $I_t$. On this basis, we adopt a simple yet effective way for ranking by using:
\begin{equation}
   R(X_t) = \frac{1}{HW} \sum_{h,w}I_t^{(h,w)}, 
\end{equation}
which is the mean value of entropy map $I_t$. Given a ranking of scores from $R(X_t)$, hyperparameter $\lambda$ is introduced as a ratio to separate the target images into an easy and hard split. Let $X_{te}$ and $X_{th}$ denote a target image assigned to the easy and hard split, respectively. In order to conduct domain separation, we define $\lambda = \frac{|X_{te}|}{|X_t|}$, where $|X_{te}|$ is the cardinality of the easy split, and $|X_{t}|$ is the cardinality of the whole target image set. To access the influence of $\lambda$, we conduct an ablation study on how to optimize $\lambda$ in Table~\ref{tab:lambda}. Note that we do not introduce a hyperparameter as the threshold value for separation. The reason is that the threshold value is dependent on a specific dataset. However, we choose a hyperparameter as ratio, which shows strong generalization to other datasets.  

%---------------------------------------------------Intra-domain adaptation -------------------------------------------------

\subsection{Intra-domain Adaptation}
Since no annotation is available for the easy split, directly aligning the gap between the easy and hard split is infeasible. But we propose to utilize the predictions from $G_{inter}$ as pseudo labels. Given an image from the easy split $X_{te}$, we forward $X_{te}$ to $G_{inter}$ and obtain the prediction map $P_{te}=G_{inter}(X_{te})$. While $P_{te}$ is a ``soft-segmentation map", we convert $P_{te}$ to $\mathcal{P}_{te}$ where each entry is a one-hot vector. With the aid of pseudo labels, $G_{intra}$ is optimized by minimizing the cross-entropy loss:    
\begin{equation}
    \begin{split}
        \mathcal{L}_{intra}^{seg}&(X_{te})  = \\
             & -\sum_{h,w}\sum_{c} \mathcal{P}_{te}^{(h,w,c)} \log \left( G_{intra}(X_{te})^{(h,w,c)} \right).
    \end{split}
\end{equation}
To bridge the intra-domain gap between easy and hard split, we adopt the alignment on the entropy map for both splits. An image $X_{th}$ from hard split  is taken as input to the generator $G$ to generate the segmentation map $P_{th}=G(X_{th})$ and the entropy map $I_{th}$. To close the intra-domain gap, the intra-domain discriminator $D_{intra}$ is trained to predict the split labels of $I_{te}$ and $I_{th}$: $I_{te}$ is from the easy split, and $I_{th}$ is from the hard split. $G$ is trained to fool $D_{intra}$. The adversarial learning loss to optimize $G_{intra}$ and $D_{intra}$ is formulated as:
\begin{equation}
    \begin{split}
    \mathcal{L}_{intra}^{adv} (X_{te}, X_{th}) =  \sum_{h,w}  \log(1 - D_{intra}(I_{th}^{(h,w)})) & \\
                                +   \log(D_{intra}(I_{te}^{(h,w)})) & .
    \end{split}
\end{equation}
Finally, our complete loss function $\mathcal{L}$ is formed by all the loss functions:
\begin{equation}
    \begin{split}
        \mathcal{L} = \mathcal{L}_{inter}^{seg}+ \mathcal{L}_{inter}^{adv} + \mathcal{L}_{intra}^{seg} +\mathcal{L}_{intra}^{adv},
    \end{split}    
\end{equation}
and our objective is to learn a target model $G$ according to:  
\begin{equation}
    \begin{split}
    G^{*} = \argmin_{G_{intra}} \min_{\mycom{G_{inter}}{G_{intra}}} \max_{\mycom{D_{inter}}{D_{intra}}}  \mathcal{L}.
    \end{split}
\end{equation}
Since our proposed model is two-step self-supervised approach, it is difficult to minimize $\mathcal{L}$ in one training stage. Thus, we choose to minimize it in three stages. First, we train the inter-domain adaptation for the model to optimize $G_{inter}$ and $D_{inter}$. Second, we generate target pseudo labels by utilizing $G_{inter}$ and rank all target images based on $S(X_t)$. Finally, we train the intra-domain adaptation to optimize $G_{intra}$ and $D_{intra}$.

% -------------------------------------------- Table GTA  --------------------------------------------------
\begin{table*}[ht!]

\begin{center}
\resizebox{\textwidth}{!}{
\begin{tabular}{l|c c c c c c c c c c c c c c c c c c c|c}
\multicolumn{21}{c}{ (a) GTA5 $\to$ Cityscapes}\\
% \hline \hline
\toprule[1.0pt]
Method & \rotatebox{90}{road} & \rotatebox{90}{sidewalk} & \rotatebox{90}{building} & \rotatebox{90}{wall} & \rotatebox{90}{fence} & \rotatebox{90}{pole} & \rotatebox{90}{light} & \rotatebox{90}{sign} & \rotatebox{90}{veg} & \rotatebox{90}{terrain} & \rotatebox{90}{sky} & \rotatebox{90}{person} & \rotatebox{90}{rider} & \rotatebox{90}{car}& \rotatebox{90}{truck} & \rotatebox{90}{bus} & \rotatebox{90}{train} & \rotatebox{90}{mbike} & \rotatebox{90}{bike} & mIoU \\
\hline
Without adaptation~\cite{tsai2018learning}  & 75.8 & 16.8 & 77.2 & 12.5 & 21.0 & 25.5 & 30.1 & 20.1 & 81.3 & 24.6 & 70.3 & 53.8 & 26.4 & 49.9 & 17.2 & 25.9 & \textbf{6.5} & 25.3 & 36.0 & 36.6 \\
ROAD~\cite{chen2018road}                    & 76.3 & 36.1 & 69.6 & 28.6 & 22.4 & \textbf{28.6} & 29.3 & 14.8 & 82.3 & 35.3 & 72.9 & 54.4 & 17.8 & 78.9 & 27.7 & 30.3 & 4.0 & 24.9 & 12.6 & 39.4 \\
AdaptSegNet~\cite{tsai2018learning}         & 86.5 & 36.0 & 79.9 & 23.4 & 23.3 & 23.9 & \textbf{35.2} & 14.8 & 83.4 & 33.3 & 75.6 & 58.5 & 27.6 & 73.7 & 32.5 & 35.4 & 3.9 & 30.1 & 28.1 & 42.4 \\
MinEnt~\cite{vu2019advent}                  & 84.2 & 25.2 & 77.0 & 17.0 & 23.3 & 24.2 & 33.3 & \textbf{26.4} & 80.7 & 32.1 & 78.7 & 57.5 & \textbf{30.0} & 77.0 & \textbf{37.9} & 44.3 & 1.8 & 31.4 & 36.9 & 43.1 \\
AdvEnt~\cite{vu2019advent}                  & 89.9 & \textbf{36.5} & 81.6 & 29.2 & \textbf{25.2} & 28.5 & 32.3 & 22.4 & 83.9 & 34.0 & 77.1 & 57.4 & 27.9 & 83.7 & 29.4 & 39.1 & 1.5 & 28.4 & 23.3 & 43.8 \\
\rowcolor{Gray} Ours                                        & \textbf{90.6} & 36.1 & \textbf{82.6} & \textbf{29.5} & 21.3 & 27.6 & 31.4 & 23.1 & \textbf{85.2} & \textbf{39.3} & \textbf{80.2} & \textbf{59.3} & 29.4 & \textbf{86.4} & 33.6 & \textbf{53.9} & 0.0 & \textbf{32.7} & \textbf{37.6} & \textbf{46.3} \\
\bottomrule
\end{tabular}}

% \vfill
\vfill 
\resizebox{\textwidth}{!}{
\begin{tabular}{l|c c c c c c c c c c c c c c c c|c|c}
\multicolumn{19}{c}{ (b) SYNTHIA $\to$ Cityscapes}\\
% \hline \hline
\toprule[1.0pt]
Method & \rotatebox{90}{road} & \rotatebox{90}{sidewalk} & \rotatebox{90}{building} & \rotatebox{90}{wall$^{*}$} & \rotatebox{90}{fence$^{*}$} & \rotatebox{90}{pole$^{*}$} & \rotatebox{90}{light} & \rotatebox{90}{sign} & \rotatebox{90}{veg} & \rotatebox{90}{sky} & \rotatebox{90}{person} & \rotatebox{90}{rider} & \rotatebox{90}{car}&  \rotatebox{90}{bus} & \rotatebox{90}{mbike} & \rotatebox{90}{bike} & mIoU & mIoU$^{*}$\\
\hline
Without adaptation~\cite{tsai2018learning}  & 55.6 & 23.8 & 74.6 & 9.2  & 0.2 & 24.4 & 6.1 & \textbf{12.1} & 74.8 & 79.0 & 55.3 & 19.1 & 39.6 & 23.3 & 13.7 & 25.0 & 33.5 & 38.6 \\
AdaptSegNet~\cite{tsai2018learning}         & 81.7 & 39.1 & 78.4 & \textbf{11.1} & 0.3 & 25.8 & 6.8 & 9.0  & 79.1 & 80.8 & 54.8 & 21.0 & 66.8 & 34.7 & 13.8 & 29.9 & 39.6 & 45.8 \\
MinEnt~\cite{vu2019advent}                  & 73.5 & 29.2 & 77.1 &  7.7 & 0.2 & \textbf{27.0} & 7.1 & 11.4 & 76.7 & 82.1 & \textbf{57.2} & 21.3 & 69.4 & 29.2 & 12.9 & 27.9 & 38.1 & 44.2  \\
AdvEnt~\cite{vu2019advent}                  & \textbf{87.0} & \textbf{44.1} & \textbf{79.7} & 9.6 & \textbf{0.6} & 24.3 & 4.8 & 7.2 & \textbf{80.1} & 83.6 & 56.4 & \textbf{23.7} & 72.7 & 32.6 & 12.8 & 33.7 & 40.8 & 47.6  \\
\rowcolor{Gray}Ours                                        & 84.3 & 37.7 & 79.5 & 5.3 & 0.4 & 24.9 & \textbf{9.2} & 8.4 & 80.0 & \textbf{84.1} & \textbf{57.2} & 23.0 & \textbf{78.0}& \textbf{38.1} & \textbf{20.3} & \textbf{36.5} & \textbf{41.7} & \textbf{48.9} \\
\bottomrule
\end{tabular}}

\vfill
\resizebox{\textwidth}{!}{
\begin{tabular}{l|c c c c c c c c c c c c c c c c c c c|c}
\multicolumn{21}{c}{ (c) Synscapes $\to$ Cityscapes}\\
% \hline \hline
\toprule[1.0pt]
Method & \rotatebox{90}{road} & \rotatebox{90}{sidewalk} & \rotatebox{90}{building} & \rotatebox{90}{wall} & \rotatebox{90}{fence} & \rotatebox{90}{pole} & \rotatebox{90}{light} & \rotatebox{90}{sign} & \rotatebox{90}{veg} & \rotatebox{90}{terrain} & \rotatebox{90}{sky} & \rotatebox{90}{person} & \rotatebox{90}{rider} & \rotatebox{90}{car}& \rotatebox{90}{truck} & \rotatebox{90}{bus} & \rotatebox{90}{train} & \rotatebox{90}{mbike} & \rotatebox{90}{bike} & mIoU \\
\hline
Without adaptation                                     & 81.8 & 40.6 & 76.1 & 23.3 & 16.8 & 36.9 & 36.8 & 40.1 & 83.0 & \textbf{34.8} & 84.9 & 59.9 & 37.7 & 78.4 & 20.4 & 20.5 & 7.8 & 27.3 & 52.5 & 45.3 \\
AdaptSegNet~\cite{tsai2018learning}      & \textbf{94.2} & \textbf{60.9} & \textbf{85.1} & 29.1 & 25.2 & \textbf{38.6} & \textbf{43.9} & 40.8 & 85.2 & 29.7 & \textbf{88.2} & 64.4 & 40.6 & \textbf{85.8} & \textbf{31.5} & 43.0 & 28.3 & 30.5 & 56.7 & 52.7 \\
\rowcolor{Gray}Ours                                     & 94.0 & 60.0 & 84.9 & \textbf{29.5}& \textbf{26.2} & 38.5 & 41.6 & \textbf{43.7} & \textbf{85.3} & 31.7 & \textbf{88.2} & \textbf{66.3} & \textbf{44.7} & 85.7 & 30.7 & 
\textbf{53.0} & \textbf{29.5} & \textbf{36.5} & \textbf{60.2} & \textbf{54.2} \\
\bottomrule
\end{tabular}}

\end{center}
\caption{The semantic segmentation results of Cityscapes validation set with models trained on GTA5 (a), SYNTHIA (b), and Synscapes (c). All the results are generated from the ResNet-101-based models. In the experiments of (a) and (b), AdvEnt\cite{vu2019advent} is used as the framework for the inter-domain adaptation and intra-domain adaptation. In the experiment of (c), AdaptSegNet~\cite{tsai2018learning} is used as the framework of the inter-domain adaptation and intra-domain adaptation. mIoU$^{*}$ in (b) denotes the mean IoU of 13 classes, excluding the classes with $^{*}$.}
\label{table:seg}
\end{table*}

\section{Experiments}
In this section, we introduce the experimental details of the inter-domain and the intra-domain adaptation on semantic segmentation. 
\subsection{Datasets}
In the experiments of semantic segmentation, we adopt the setting of adaptation from the synthetic to the real domain. To conduct this series of tests, synthetic datasets including GTA5~\cite{Richter_2016_ECCV}, SYNTHIA~\cite{ros2016synthia} and Synscapes~\cite{wrenninge2018synscapes} are used as source domains, along with the real-world dataset Cityscapes~\cite{Cordts2016Cityscapes} as the target domain. Models are trained given labeled source data and unlabeled target data. Our model is evaluated on Cityscapes validation set. 

\begin{itemize}
    \item GTA5: The synthetic dataset GTA5~\cite{Richter_2016_ECCV} contains 24,966 synthetic images with a resolution of 1,914$\times$1,052 and corresponding ground-truth annotations. These synthetic images are collected from a video game based on the urban scenery of Los Angeles city. The ground-truth annotations generated automatically contain 33 categories. For training, we consider only 19 categories which are compatible with the Cityscapes dataset~\cite{Cordts2016Cityscapes}, similarly to previous work.
    \item SYNTHIA: SYNTHIA-RAND-CITYSCAPES~\cite{ros2016synthia} is used as another synthetic dataset. It contains 9,400 fully annotated RGB images. During the training time, we consider the 16 common categories with the Cityscapes dataset. During evaluation, 16- and 13- class subsets are used to evaluate the performance.
    \item Synscapes: Synscapes~\cite{wrenninge2018synscapes} is a photorealistic synthetic dataset consisting of 25,000 fully annotated RGB images with a resolution of 1,440$\times$720. Alike Cityscape, the ground-truth annotations contain 19 categories.
    \item Cityscapes: As the dataset collected from real world, Cityscapes~\cite{Cordts2016Cityscapes} provides 3,975 images with fine segmentation annotations. 2,975 images are taken from the training set of Cityscapes to be used for training. The 500 images from the evaluation set of Cityscapes are used to evaluate the performance of our model.
\end{itemize}

\noindent \textbf{Evaluation.} The semantic segmentation performance is evaluated on every category using the PASCAL VOC intersection-over-union metric, \ie, $\text{IoU} = {\text{TP}}/{(\text{TP}+\text{FP}+\text{FN})}$~\cite{everingham2015pascal}, where TP, FP, and FN are the number of true positive, false positive, and false negative pixels, respectively. \\

\noindent \textbf{Implementation Details.} In the experiments for GTA5$\to$Cityscapes and SYNTHIA$\to$Cityscapes, we utilize the framework of AdvEnt~\cite{vu2019advent} to train $G_{inter}$ and $D_{inter}$ for inter-domain adaptation; the backbone of $G_{inter}$ is a ResNet-101 architecture~\cite{he2016deep} with pretrained parameters from ImageNet~\cite{deng2009imagenet}; the input data are labeled source images and unlabeled target images. The model for inter-domain adaptation $G_{inter}$ is trained for 70,000 iterations. After training, $G_{inter}$ is used to generate the segmentation and entropy maps for all 2,975 images from Cityscapes training set. Then, we utilize $R(X_t)$ to get the ranking scores for all target images and to separate them into the easy and hard split based on $\lambda$. We conduct an ablation study of $\lambda$ for optimization in Table~\ref{tab:lambda}. For the intra-domain adaptation, $G_{intra}$ has same architecture as $G_{inter}$, and $D_{intra}$ same as $D_{inter}$; the input data are 2,975 Cityscapes training images with pseudo labels of easy split. $G_{intra}$ is trained with the pretrained parameters from ImageNet and $D_{intra}$ from scratch, similar to AdvEnt. In addition to the previously mentioned experiments, we also conduct the experiment for Synscapes$\to$Cityscapes. For comparison with AdaptSegNet~\cite{tsai2018learning}, We apply the framework of AdaptSegNet in the experiment for the inter-domain and intra-domain adaptation.

Similarly to~\cite{vu2019advent} and~\cite{tsai2018learning},  we utilize the multi-level feature outputs from \textit{conv4} and \textit{conv5} for inter-domain adaptation and intra-domain adaptation. To train $G_{inter}$ and $G_{intra}$, we apply an SGD optimizer~\cite{bottou2010large} with a learning rate of $2.5\times 10^{-4}$, momentum $0.9$, and a weight decay $10^{-4}$ for training $G_{inter}$ and $G_{intra}$. An Adam optimizer~\cite{kingma2014adam} with a learning rate of $10^{-4}$ is used for training $D_{inter}$ and $D_{intra}$.

%------------------------------------------------ Figure ------------------------------------------------------------
\begin{figure*}
    \centering 
    \includegraphics[width=0.98\textwidth]{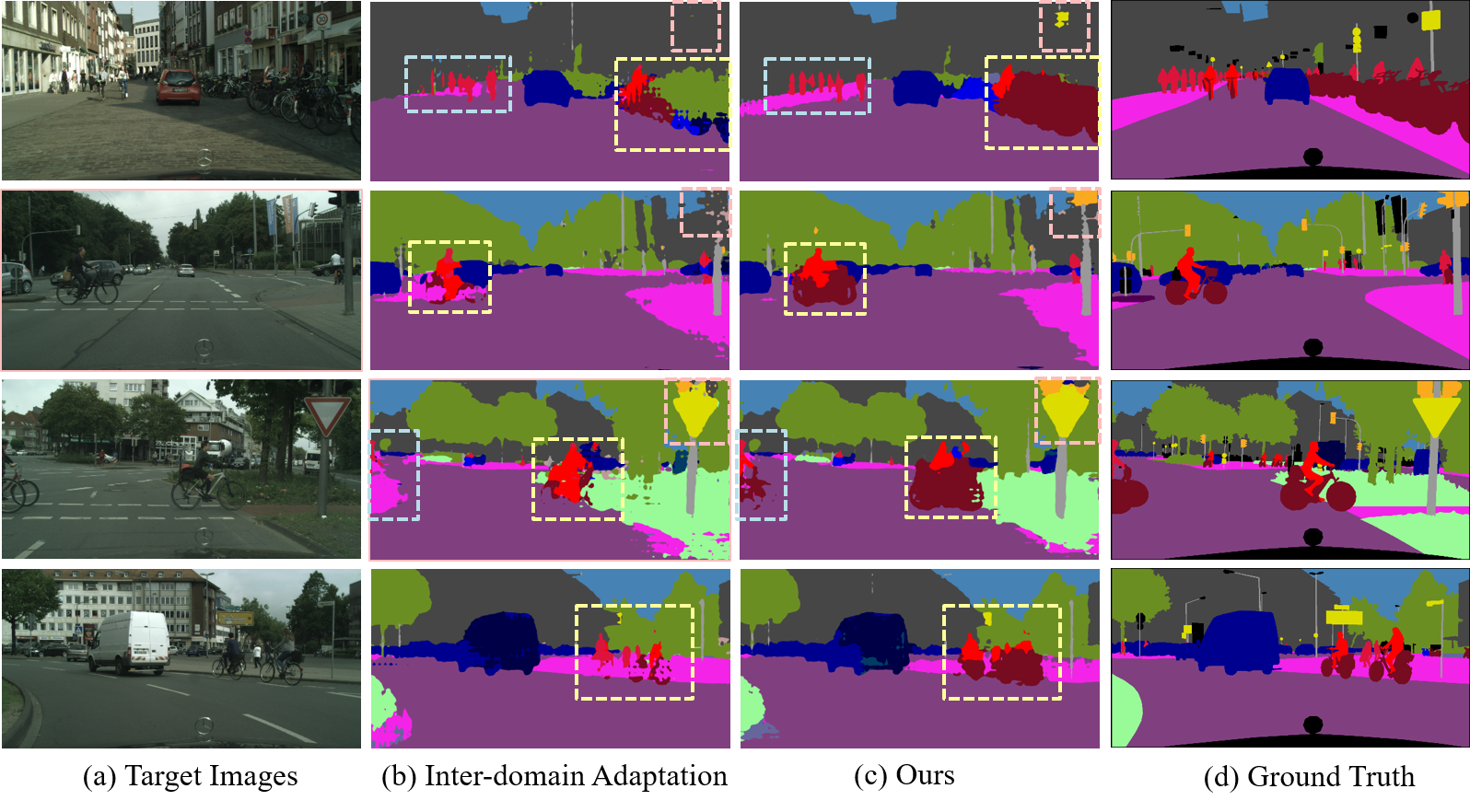}
\caption{The example results of evaluation for GTA5$\to$Cityscapes. (a) and (d) are the images from Cityscapes validation set and the corresponding ground-truth annotation. (b) are the predicted segmentation maps of the inter-domain adaptation~\cite{vu2019advent}. (c) are the predicted maps from our technique.}
    \label{fig:target_images}
\end{figure*}

\begin{figure*}
    \centering 
    \includegraphics[width=\textwidth]{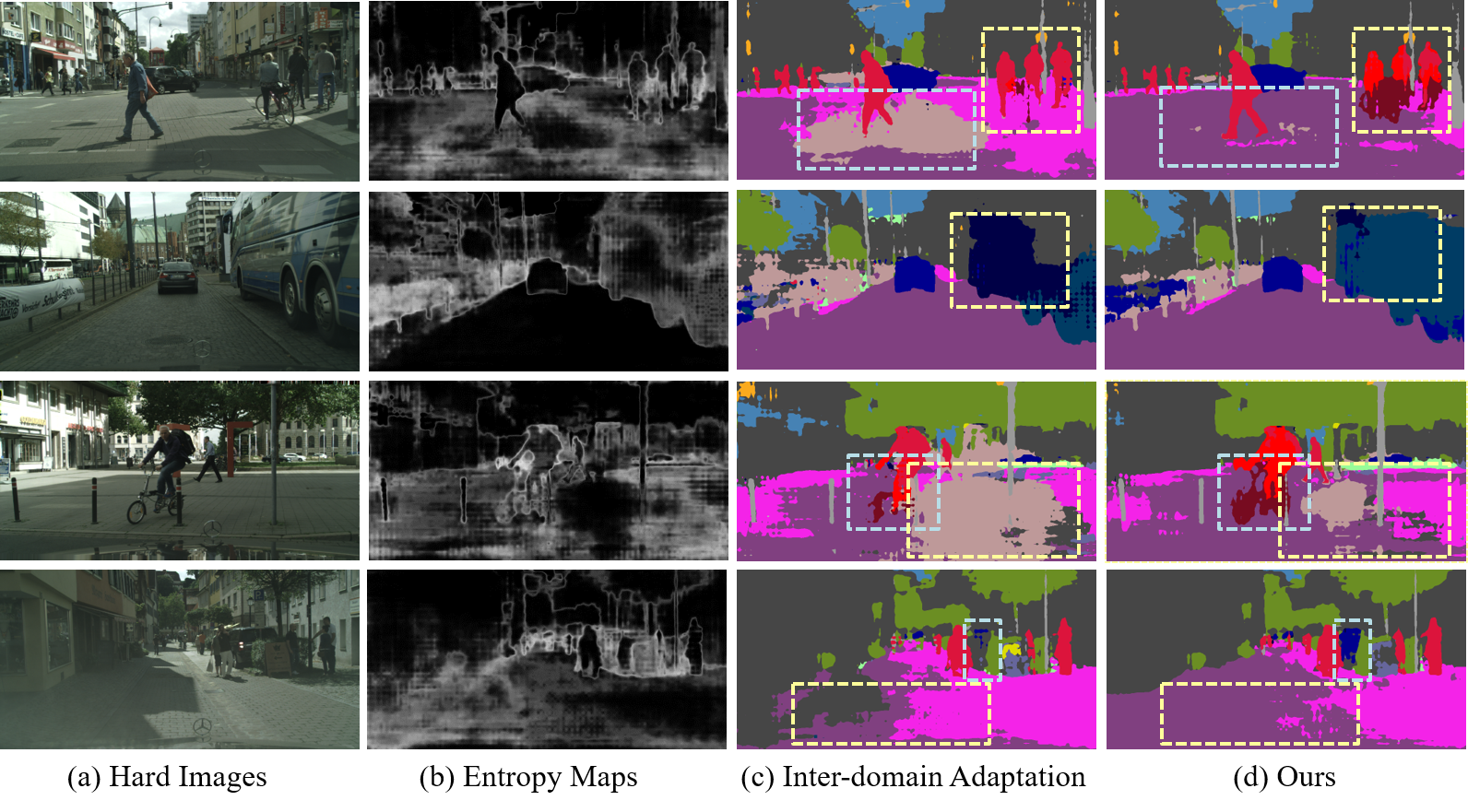}
    \caption{The examples of entropy maps from hard split for GTA5$\to$Cityscapes. (a) are the hard images from Cityscapes training set. (b) and (c) are the predicted entropy and segmentation maps from model trained solely by the inter-domain adaptation~\cite{vu2019advent}. (d) are the improved predicted segmentation results of the hard images from our model.}
    \label{fig:easy_hard}
\end{figure*}

% $G_{inter}$ serves as a classifier and the architecture is based on a variant of the LeNet architecture. 
% ----------------------------- table ablation study lambda ----------------------------------

\begin{table}[t!]
    \centering
    \resizebox{0.85\textwidth}{!}{
    \begin{tabular}{c|c c c c c c}
    \toprule[1.0pt]
    \multicolumn{7}{c}{GTA5 $\to$ Cityscapes}\\
    \hline
    $\lambda$ & 0.0 & 0.5 & 0.6 & 0.67 & 0.7 & 1.0 \\
    % \hline
    % ${\#X_T^{e}}/{\#X_T}$ & $8/15$ & ${9}/{15}$ & ${10}/{15}$ & ${11}/{15}$\\
    \hline
    mIoU & 43.8 & 45.2 &  46.0  & \textbf{46.3}  & 45.6 & 45.5 \\
    \bottomrule
    \end{tabular}}
    \caption{The ablation study on hyperparameter $\lambda$ for separating the target domain into the easy and the hard split.}
    \label{tab:lambda}
\end{table}

% ------------------------------------- table self-train vs. intra-domain adaptation -------------------------------------
\begin{table}[t!]
    \centering
    \resizebox{0.82\linewidth}{!}{
    \begin{tabular}{l c c}
    \toprule[1.0pt]
    Model & & mIoU  \\
    \hline
    AdvEnt [\textcolor{green}{25}] &  & 43.8 \\
    AdvEnt + intra-domain adaptation &  &  45.1 \\
    AdvEnt + self-training $(\lambda=1.0)$ &  & 45.5   \\
    \hline 
    Ours &  &  46.3 \\
    Ours + {\small entropy normalization} & & \textbf{47.0} \\
    \bottomrule
    % \hline
    \end{tabular}
    }
    \caption{The self-training and intra-domain adaptation gain on GTA5 $\to$ Cityscapes.}
    \label{tab:self-train}
\end{table}

% ------------------------------------- classification -------------------------------------

\subsection{Results}

\noindent \textbf{GTA5.} In Table~\ref{table:seg} (a), we compare the segmentation performance of our method with other state-of-the-art methods~\cite{tsai2018learning, chen2018road, vu2019advent} on Cityscapes validation set. For a fair comparison, the baseline model is adopted from DeepLab-v2\cite{chen2017deeplab} with a ResNet-101 backbone. Overall, our proposed method achieves $46.3\%$ in mean IoU. Compared to AdvEnt, the intra-domain adaptation from our method leads to a $2.5\%$ improvement in mean IoU. 

To highlight the relevance of the proposed intra-domain adaptation, we conduct a comparison with segmentation loss $\mathcal{L}_{intra}^{seg}$ and adversarial adaptation loss $\mathcal{L}_{intra}^{adv}$ in Table~\ref{tab:self-train}. The baseline AdvEnt~\cite{vu2019advent} achieves $43.8\%$ of mIoU. By using AdvEnt + intra-domain adaptation, which means $\mathcal{L}_{intra}^{seg}=0$, we obtain $45.1\%$, showing the effectiveness of adversarial learning for the intra-domain alignment. By applying AdvEnt + self-training, with $\lambda=1.0$ (all pseudo labels used for self-training), which means $\mathcal{L}_{intra}^{adv}=0$, we achieve $45.5\%$ of mIoU, underlying the importance of employing pseudo labels. Finally, our proposed model achieves $46.3\%$ of mIOU (self-training + intra-domain alignment).

Admittedly, complex scenes (containing many objects) might be categorized as ``hard". To provide a more representative ``ranking", we adopt a new normalization by dividing the mean entropy with the number of predicted rare classes in the target image. For Cityscapes dataset, we define these rare classes as ``wall, fence, pole, traffic light, traffic sign, terrain, rider, truck, bus, train, motor". The entropy normalization helps to move images with many objects to the easy split. By using the normalization, our proposed model achieves $47.0\%$ of mIoU, as shown in Table~\ref{tab:self-train}. Our proposed method also has limitation for some classes. 
% In Table~\ref{table:seg}(a), the IoU of ``train" class from our method is 0.0, lower than other existing methods. The reason behind is that, pixels from the ``train" samples are scarcely distributed in the easy split.
% The idea of the intra-domain adaptation in our method is to utilize the pseudo labels of the easy samples to improve the prediction results of hard samples. Therefore, the ``train" class are easily merged with background classes by the intra-domain adaptation.

In Figure~\ref{fig:target_images}, we provide some visualizations of segmentation maps from our technique. The segmentation maps generated from our model trained with inter-domain alignment and intra-domain alignment are more accurate than the baseline model AdvEnt, which has been only trained with inter-domain alignment. A representative set of images belonging to the ``hard" split are visible in Figure~\ref{fig:easy_hard}. After intra-domain alignment, we produce the segmentation maps shown in the (d) column. Compared with (c) column, our model can be transferred to more difficult target images. \\

\noindent \textbf{Analysis of Hyperparameter $\lambda$.} We conduct a study on finding a proper value for the hyperparameter $\lambda$ in our experiment of GTA5$\to$Cityscapes. In Table \ref{tab:lambda}, different values of $\delta$ are used for setting up the decision boundary for domain separation. When $\lambda=0.67$, \ie., the ratio of $|X_{te}|$ to $|X_t|$ is approximately $2/3$, the model achieves $46.3$ of mIoU as the best performance on Cityscapes validation set. \\

\noindent \textbf{SYNTHIA.} We use SYNTHIA as the source domain and present evaluation results of the proposed method and state-of-the-art methods~\cite{tsai2018learning, vu2019advent} on Cityscapes validation set in the Table \ref{table:seg}. For a fair comparison, we also adopt the same DeepLab-v2 with the ResNet-101 architecture. Our method is evaluated on both 16-class and 13-class baselines. According to the results in Table~\ref{table:seg} (b), our proposed method has achieved $41.7\%$ and $48.9\%$ of mean IoU on 16-class and 13-class baseline, respectively. As shown in the Table \ref{table:seg}, our model is significantly more accurate on the car and motor bike classes than existing techniques. The reason is that we apply the intra-domain adaptation to further narrow the domain gap.\\

\noindent \textbf{Synscapes.} The only work that we currently have found using Syncapes dataset is ~\cite{tsai2018learning}. Thus we use AdaptSegNet~\cite{tsai2018learning} as our baseline model. In order to present a fair comparison, We only consider using vanilla-GAN in our experiments. With inter-domain and intra-domain adaptation, our model achieves $54.2\%$ of mIoU, which higher than AdaptSegNet shown in Table~\ref{table:seg} (c).

\subsection{Discussion}

\noindent \textbf{Theoretical Analysis.} Comparing (a), (b) in Table~\ref{table:seg}, GTA5 to Cityscapes is more effective than SYNTHIA to Cityscapes. We believe the reason is that GTA5 has more similar images of street scenes with Cityscapes than other synthetic datasets. We also provide a theoretical analysis here. Let $\mathcal{H}$ denote the hypothesis class, $S$ and $T$ be the source and the target domain. The theory from ~\cite{ben2007analysis} proposes to bound the expected error on the target domain $\epsilon_T(h)$: $\forall h \in \mathcal{H}, \epsilon_T(h) \leq \epsilon_S(h) + \frac{1}{2}d_{\mathcal{H}}(S,T) + \Lambda$, where $\epsilon_S(h)$ is the expected error on the source domain; $d_{\mathcal{H}}(S,T)=2\sup|\text{Pr}_{S}(h)-\text{Pr}_{T}(h)|$, which is the distance for domain divergence; $\Lambda$ is considered as a constant in normal cases. Therefore, $\epsilon_T(h)$ is upper bounded by $\epsilon_S(h)$ and $d_{\mathcal{H}}(S,T)$ in our case. Our proposed model is to minimize $d_{\mathcal{H}}(S,T)$ by using the inter-domain and the intra-domain alignment together. If $d_{\mathcal{H}}(S,T)$ has high value, the higher upper bound in the first stage of the inter-domain adaptation affects our entropy ranking system, and the intra-domain adaptation processes. Therefore, our model is less efficient in big domain gap. With respect to the limitation, our model performance is affected by $d_{\mathcal{H}}(S,T)$ and $\epsilon_S(h)$. Firstly, the larger divergence of the source and target domain leads to higher value in $d_{\mathcal{H}}(S,T)$. The upper bound of error is higher so our model would be less effective. Secondly, $\epsilon_S(h)$ would be very high when the model uses small neural networks. In such case, our model would also be less effective.\\

\begin{table}[t!]
    \centering
    \resizebox{1.0\linewidth}{!}
    {
    \begin{tabular}{l c c c}
    \toprule[1.0pt]
    Model & MNIST $\to$ USPS  & USPS $\to$ MNIST  & SVHN $\to$ MNIST  \\
    \hline
    Source only       & 82.2$\pm$ 0.8 & 69.6$\pm$ 3.8 & 67.1$\pm$ 0.6 \\
    ADDA~\cite{tzeng2017adversarial}              & 89.4$\pm$ 0.2 & 90.1$\pm$ 0.8 & 76.0$\pm$ 1.8 \\
    CyCADA~\cite{pmlr-v80-hoffman18a}            & 95.6$\pm$ 0.2 & 96.5$\pm$ 0.1 & 90.4$\pm$ 0.4 \\
    Ours              & \textbf{95.8}$\pm$\textbf{0.1}        & \textbf{97.8}$\pm$\textbf{0.1}        & \textbf{95.1}$\pm$\textbf{0.3}        \\
    \bottomrule
    \end{tabular}
    }
    \caption{The experimental results of adaptation across digit datasets.}
    \label{tab:lambda3}
\end{table}

\noindent \textbf{Digit Classification.} Our model are also capable to be applied in digit classification task. We consider the adaptation shift of MNIST$\to$USPS,  USPS$\to$MNIST, and SVHN$\to$MNIST. Our model is trained using the training sets: MNIST with 60,000 images, USPS with 7,291 images, standard SVHN with 73,257 images. The proposed model is evaluated on the standard test sets: MNIST with 10,000 images and USPS with 2,007 images. In digit classification task, $G_{inter}$ and $G_{intra}$ serve as classifiers with same architecture, which is based on a variant of the LeNet architecture. In inter-domain adaptation, We utilize the framework of CyCADA~\cite{pmlr-v80-hoffman18a} to train $G_{inter}$ and $D_{inter}$. In the ranking stage, we utilize $G_{inter}$ to generate the predictions of all target data and compute their ranking score using $R(X_t)$. With respect to $\lambda$, we adopted $\lambda=0.8$ in all experiments. Our network for intra-domain adaptation is also based on CyCADA~\cite{pmlr-v80-hoffman18a}. In Table~\ref{tab:lambda3}, our proposed model achieve 95.8$\pm$0.1$\%$ of accuracy on MNIST $\to$ USPS, 97.8$\pm$0.1$\%$ of accuracy on USPS $\to$ MNIST, and 95.1$\pm$0.3$\%$ on SVHN $\to$ MNIST. Our model outperforms the baseline model CyCADA~\cite{pmlr-v80-hoffman18a}.

\section{Conclusion}
In this paper, we present a self-supervised domain adaptation to minimize the inter-domain and intra-domain gap simultaneously. We first train the model using the inter-domain adaptation from existing approaches. Secondly, we produce target image entropy maps and use an entropy-based ranking functions to split the target domain. Lastly, we conduct the intra-domain adaptation to further narrow the domain gap. We conduct extensive experiments on synthetic to real images in traffic scenarios. Our model can be combined with existing domain adaptation approaches. Experimental results shows that our model outperforms existing adaptation algorithms. 

\section*{Acknowledgments}
This research was partially supported by the Shared Sensing for Cooperative Cars Project funded by Bosch (China) Investment  Ltd.  This  work  was  also  partially supported  by the Korea Research Fellowship Program through the National Research Foundation of Korea (NRF) funded by the Ministry of Science, ICT and Future Planning (2015H1D3A1066564).

{\small
\bibliographystyle{ieee_fullname}
\bibliography{egbib}
}

\end{document}